\documentclass[11pt]{article}
\usepackage[a4paper, margin=1in]{geometry}
\usepackage[numbers]{natbib}
\usepackage{amsmath}
\usepackage{algorithm}
\usepackage{booktabs} 
\usepackage{algorithmic}
\usepackage{multirow}
\usepackage[table]{xcolor}
\usepackage{amsmath}
\usepackage{array}

\title{HyperQ-Opt: Q-learning for Hyperparameter Optimization}
\author{\small Md. Tarek Hasan\\\small Department of CSE, United International University\\\small United City, Madani Ave, Dhaka 1212\\\small Email: tarek@cse.uiu.ac.bd}
\date{}

\begin{document}

\maketitle

\begin{abstract}

Hyperparameter optimization (HPO) is critical for enhancing the performance of machine learning models, yet it often involves a computationally intensive search across a large parameter space. Traditional approaches such as Grid Search and Random Search suffer from inefficiency and limited scalability, while surrogate models like Sequential Model-based Bayesian Optimization (SMBO) rely heavily on heuristic predictions that can lead to suboptimal results. This paper presents a novel perspective on HPO by formulating it as a sequential decision-making problem and leveraging Q-learning, a reinforcement learning technique, to optimize hyperparameters. The study explores the works of H.S. Jomaa et al. and Qi et al., which model HPO as a Markov Decision Process (MDP) and utilize Q-learning to iteratively refine hyperparameter settings. The approaches are evaluated for their ability to find optimal or near-optimal configurations within a limited number of trials, demonstrating the potential of reinforcement learning to outperform conventional methods. Additionally, this paper identifies research gaps in existing formulations, including the limitations of discrete search spaces and reliance on heuristic policies, and suggests avenues for future exploration. By shifting the paradigm toward policy-based optimization, this work contributes to advancing HPO methods for scalable and efficient machine learning applications. 

\end{abstract}

\section{Introduction}
Selecting the right combination of hyperparameters is crucial for a machine learning-based model to perform to its full potential. Hyperparameter Optimization (HPO) or finding the optimal setting of hyperparameters for which we get the maximum expected outcome is an open-ended research field going through the exploration phase and piloting different strategies. Manual Search or tuning the values of hyperparameters based on cognitive memory is a typical approach to figuring out the optimal combination of hyperparameters. However, it requires in-depth knowledge of the domain, model, and dataset which works like a hurdle for the researchers. That is why we need to automate the process of finding optimal combinations of hyperparameters using the power of Artificial Intelligence. One automated approach that is typically used for HPO is Grid Search \cite{bergstra2011algorithms}.\\

To begin with, let's define the problem HPO formally. Let $M$ represent a machine learning algorithm having $N$ hyperparameters. We define $\Lambda_n$ as the domain of the $n^{th}$ hyperparameter and $\Lambda=\Lambda_1 \times \Lambda_2 \times \ldots \Lambda_N$ as the whole hyperparameter setting space, or we can define it as search space. $M_\lambda$ \cite{hutter2019automated} represents $M$ with its hyperparameter setting instantiated to $\lambda$, and $\lambda \in \Lambda$ represents a vector of hyperparameter configuration. \\

The loss function computes the disparity between the current and expected outputs of the algorithm. By adjusting hyperparameters on datasets, we want to minimize the loss function of the algorithm. Hence, it can be defined as a minimization problem. Stated differently, given a dataset $D$, we want to find $$ \lambda^*=\underset{\lambda \in M}{\arg \min } 
L\left(M_\lambda, D_{\text {train }}, D_{\text {valid}}\right) $$ where the loss function that $M_\lambda$ achieves when trained on $D_{train }$ and evaluated on $D_{{valid. }}$ is $L\left(M_\lambda, D_{train }, D_{valid }\right)$. For the sake of example, we use the loss function as the performance metric here, but we can use any rational performance metric, and the nature of the optimization changes accordingly. For instance, it transforms into a maximization problem if accuracy is used as the metric.\\

Grid Search \cite{bergstra2011algorithms} is an automated approach to selecting the optimal set of hyperparameters from a grid of predefined possible hyperparameter values. That means it executes the model \textit{A} on all combinations of hyperparameters present in the grid and keeps track of the loss function. Finally, it chooses the combination of hyperparameters for which the loss function is minimized. However, the problem arises when the model \textit{A} is computationally expensive to execute, and then it becomes infeasible to execute the model on all possible combinations of hyperparameters. In addition, as the number of hyperparameters grows, the size of the grid becomes infeasible to train on the model. To mitigate the computation, another approach has been introduced Random Search \cite{bergstra2012random} that randomly samples hyperparameter values from predefined ranges and evaluates their performances. Using a random search may not be the most effective way to explore the search space, and obtaining the optimum hyperparameters within a specified number of epochs is not certain. The two most popular methods for hyperparameter optimization are Grid Search and Random Search. To overcome the limitation of random search, a model named Sequential Model-based Bayesian Optimization (SMBO) \cite{hutter2011sequential}.\\

SMBO \cite{hutter2011sequential} is effective for tasks such as hyperparameter tuning in machine learning because it progressively constructs a probabilistic model of the function and uses it to direct the search for the optimal solution. After building a surrogate model to calculate the loss for a specific model, dataset, and hyperparameter setup, it uses a heuristic function to sequentially choose the next hyperparameter to be evaluated. As the model is based on a heuristic function to decide the next hyperparameter setup, the hyperparameter setting might stuck on a suboptimal setting if the prediction is wrong.\\

An agent that interacts with its environment to learn to make decisions is called an agent in reinforcement learning \cite{kaelbling1996reinforcement}. Through trial and error, the agent learns how to maximize the cumulative benefits it receives from its surroundings. It requires selecting a policy, a mapping from states to actions—to accomplish long-term objectives via a sequence of decision-making tasks. HPO problem can be developed as a sequential decision-making problem, in which a set of hyperparameters will be chosen next to evaluate can be decided. Then, we can solve this problem using Reinforcement Learning which helps us not to depend on a heuristic function as SMBO.\\

The HPO problem can be stated as a sequential decision-making problem or Markov Decision Problem (MDP), where the agent selects which hyperparameter to evaluate next. Then, by applying reinforcement learning to address this problem, we can avoid using a heuristic function like SMBO. H.S. Jomaa et al. \cite{jomaa2019hyp} are the first ones to formulate the problem of HPO as the sequential decision problem, and handle it with a model-free reinforcement learning approach. To formulate the HPO problem as MDP, we need to define a tuple of the following items $(S, A, P, R)$ where \textit{S}, \textit{A}, \textit{P}, \textit{R} represents state, action, transition probabilities, and reward, sequentially. \\ 

The use of reinforcement learning techniques to solve the HPO problem has gained prominence in recent years \cite{pmlr-v130-zhang21n, talaat2023rl}, particularly in conditions involving dynamic environments, non-convex objective functions, complex search spaces, constrained computational resources, and interactions with external systems. Reinforcement learning algorithms can outperform conventional optimization techniques by learning optimal strategies for choosing hyperparameters over time by formulating HPO as a sequential decision-making problem. This paper explores the studies that formulate HPO as MDP and utilize Q-learning \cite{watkins1992q} to address HPO which can be a good contribution for the interested researcher in this field.  Q-learning is a reinforcement learning method that updates Q-values in response to rewards and transitions an agent experiences. This allows it to learn the best action-selection strategy for any finite Markov decision process. For each study, the way of MDP formulation, the proposed algorithm, and the key findings have been presented sequentially.

\section{Literature Review}
H.S. Jomaa et al. \cite{jomaa2019hyp} model the Hyperparameter Optimization (HPO) problem as a Markov Decision Process (MDP), where they define a discrete finite search space for hyperparameters by specifying a predefined list of values for each hyperparameter for the model $M$, Long Short Term Memory (LSTM) \cite{hochreiter1997long}. They categorize hyperparameters into three groups: structure, optimization, and regularization. The hyperparameter grid detailing these categories is presented in Table \ref{tab:HypGrid}. \\

\begin{table}[H]
\centering
\caption{Hyperparameter grid employed by H.S. Jomaa et al. \cite{jomaa2019hyp}}
\begin{tabular}{|l|l|l|}
\hline
\multicolumn{3}{|c|}{\cellcolor{gray!30}\textbf{Hyperparameter Grid}} \\ 
\hline
\textbf{Hyperparameter} & \textbf{Values} & \textbf{Encoding} \\
\hline
\multicolumn{3}{|c|}{\cellcolor{gray!30}\textit{Structure}} \\ 
\hline
Activation Function & ReLU, LeakyReLU, tanh & One-hot \\
\hline
Number of Neurons & 5, 10, 20 & Scalar \\
\hline
Number of Hidden Units & 10, 20, 50 & Scalar \\
\hline
\multicolumn{3}{|c|}{\cellcolor{gray!30}\textit{Optimization}} \\ 
\hline
Optimizer & Adam, AdaDelta, AdaGrad & One-hot \\
\hline
Number of Epochs & 10, 100 & Scalar \\
\hline
\multicolumn{3}{|c|}{\cellcolor{gray!30}\textit{Regularization}} \\ 
\hline
Dropout  & 0, 0.2, 0.4 & Scalar \\
\hline
Regularization Lp & L1, L2 & One-hot \\
\hline
Regularization Constant & 0.01, 0.001, 0.0001 & Scalar \\
\hline
\end{tabular}
\label{tab:HypGrid}
\end{table}

In their formulation, a state ($s_t$) comprises the meta-features of the dataset D, the current hyperparameter configuration ($\lambda_t$), and the cumulative reward ($r_t$) obtained for that configuration. High-level attributes, such as the number of instances, features, feature types, and statistical summaries, that describe the attributes and structure of a dataset are referred to as meta-features. They are used in hyperparameter optimization and meta-learning to help decision-making. Formally, the state can be represented as $s_{t} = (meta feature (D), (\lambda_t, r_t))$. The action space consists of all possible combinations of hyperparameter settings, $A$ from the predefined grid, resulting in a total number of actions equal to the cardinality of the hyperparameter grid, denoted as $\Lambda$.\\

The reward at time step $t$ ($R_t$) is defined based on the performance metric, represented by a modified loss function value. Transitions between states occur when the agent switches to a new hyperparameter setting, leading to the next state, $s_{t+1}$, which comprises the dataset's meta-features, the updated hyperparameter configuration ($\lambda'_t$), and the sum of the previous cumulative reward and the reward obtained for the current hyperparameter setting ($R_t + r_t$). Hence, $s_{t+1} = (meta feature (D), (\lambda'_t, R_t+ r_t))$.\\

To determine the termination of an episode, two conditions are employed. Firstly, a maximum number of trials is predefined, ensuring that episodes do not exceed a specified duration. Secondly, termination occurs if the same action is selected twice consecutively, encouraging exploration of the hyperparameter space by preventing the agent from becoming stuck in repetitive actions.\\

They build a Q network to predict the state-action value, or Q values, and train the parameters of the Q network rather than looking for the best hyperparameter setting for a dataset and a model. They can therefore use the Q network to determine the best hyperparameter value for any given dataset using the specified model classified as transfer learning.\\

The HPO problem is formulated as a sequential decision problem by Qi et al. \cite{qi2023hyperparameter}. They define a discrete finite search space for hyperparameter by providing a predefined list of values for each hyperparameter for the models of Bidirectional Long Short Term Memory (BiLSTM) \cite{schuster1997bidirectional} and Convolutional Neural Network (CNN) \cite{lecun1998gradient}. Table \ref{tab:QGrid} illustrates the search space (exclusively for CNN) used by Qi et al. for this demonstration. \\

\begin{table}[h]
\centering
\caption{CNN Hyperparameter Search Space \cite{qi2023hyperparameter}}
\begin{tabular}{|l|l|}
\hline
\rowcolor{gray!30}
\textbf{Hyperparameters}                           & \textbf{Values}                   \\ \hline
Learning rate                                      & \{0.0001, 0.001, 0.01, 0.1\}      \\ \hline
Momentum                                           & \{0.5, 0.9, 0.95, 0.99\}          \\ \hline
Kernel size                                        & \{1 $\times$ 1, 3 $\times$ 3, 5 $\times$ 5\} \\ \hline
Number of units in fully connected layer           & \{128, 256, 512, 1024\}           \\ \hline
Dropout rate                                       & \{0.3, 0.4, 0.5, 0.6, 0.7\}       \\ \hline
Batch size                                         & \{16, 32, 64, 128\}               \\ \hline
\end{tabular}
\label{tab:QGrid}
\end{table}

A state in their formulation consists of a configuration of hyperparameters ($\lambda_t$). For every action, only one hyperparameter is tuned to minimize the dimensionality of the action space. The reward is the difference between the accuracy score of the neural network model before hyperparameter tuning and the current one. $ACC(M, \lambda_{t}) - ACC(M, \lambda_{t-1})$ is the value of $R_t$. As defined by the transition function of this formulation, the current hyperparameter setting generated as action will be the next state. \\

There are two criteria used to decide when an episode ends. First, a predetermined maximum number of trials is used to ensure the limitation of the resources or trials. Second, termination occurs if the performance of the neural network model with the current hyperparameter setting deviates from the prior one by more than 1\%. They start each episode with the hyperparameter configuration that is currently performing the best. The ability of their method to consistently search using the present optimal hyperparameter setting is driven by the second condition of termination and the initial state setup strategy, which both increase search efficiency. Qi et al. \cite{qi2023hyperparameter} look for the best hyperparameter configuration for a particular model and a particular dataset, in contrast to H.S. Jomaa et al. \cite{jomaa2019hyp}. Thus, to determine the optimal hyperparameter configuration, they must rerun the computational method if the dataset changes.

\section{Algorithm}

\begin{algorithm}[]
\caption{Hyp-RL \cite{jomaa2019hyp} Algorithm}
\begin{algorithmic}[1]
\REQUIRE $D$ - datasets, $\Lambda$ - hyperparameter grid, $\gamma$ - discount factor, $N_u$ - target update frequency, $N_b$ - replay buffer size, $N_e$ - number of episodes per dataset, $T$ - number of actions per episode, $\epsilon$ - exploration rate
\STATE initialize $\hat{Q}$ network parameters $\theta$ randomly; $\theta^-$ = $\theta$; replay buffer $B$ = $\emptyset$
\FOR{$N_e \times |D|$ iterations}
    \STATE $s_0$ = (metafeatures($D$), ($\{0\}^{\text{dim}(\Lambda)},0$)), $D \sim \text{Unif}(D)$
    \FOR{$t \in 0,...,T$}
        \STATE $s_t =$ metafeatures($D$)
        \WHILE{$s_t$ is not terminal and $s_t \neq s_{t+1}$}
            \STATE Determine next action $a_t$:
            \STATE \quad - With probability $\epsilon$: $a_t \sim \text{Unif}(\Lambda)$
            \STATE \quad - Otherwise: $a_t = \arg\max_a \hat{Q}(s_t,a;\theta)$
            \STATE Receive reward $r_t = R(D,\lambda = a_t)$
            \STATE Generate new state $s_{t+1} = \tau(s_t,\lambda_t,r_t)$
            \IF{$s_t = s_{t+1}$}
                \STATE Terminate current episode
            \ENDIF
            \STATE Store experience in replay buffer
            \STATE $B = B \cup \{(s_t,s_{t+1},a_t,r_t)\}$
            \IF{size($B$) $> N_b$}
                \STATE Remove oldest experience from $B$
            \ENDIF
            \STATE Sample and relabel a minibatch $B$ of experiences from the replay buffer
            \STATE $minibatch = \text{sample\_and\_relabel\_minibatch}(B)$
            \FOR{$(s, s', a, r)$ in $minibatch$}
                \IF{$s'$ is not terminal}
                    \STATE $\hat{Q}_{\text{target}}(s',a, r) = r + \gamma \max_a \hat{Q}_{\text{target}}(s',a;\theta^-)$
                \ELSE
                    \STATE $\hat{Q}_{\text{target}}(s',a, r) = r$
                \ENDIF
                \STATE
                $\theta = \arg\min_{\theta'} \sum_{(s, a, Q) \in B} \left(Q - \hat{Q}(s, a ; \theta')\right)^2$
            \ENDFOR
            \IF{$t$ is multiple of $N_u$}
                \STATE $\theta^- = \theta$
            \ENDIF
            \STATE $t = t + 1$
        \ENDWHILE
    \ENDFOR
\ENDFOR
\RETURN $\theta$
\end{algorithmic}
\label{algo:Hyp-RL}
\end{algorithm}

The algorithm designed by H.S. Jomaa et al. \cite{jomaa2019hyp} named as Hyp-RL shown in Algorithm \ref{algo:Hyp-RL} takes several parameters as input, including datasets ($D$), a hyperparameter grid ($\Lambda$), a discount factor ($\gamma$), target update frequency ($N_u$), replay buffer size ($N_b$), number of episodes per dataset ($N_e$), number of actions per episode ($T$), and an exploration rate ($\epsilon$).\\

The algorithm initializes the parameters of the Q-network ($\hat{Q}$) randomly and sets the target Q-network parameters ($\theta^-$) to be the same as the initial Q-network parameters. It also initializes an empty replay buffer ($B$).\\ 

The algorithm then iterates over the total number of episodes ($N_e \times |D|$). For each episode, it initializes the initial state ($s_0$) by obtaining meta-features of the dataset ($D$) and sets the hyperparameter configuration to an initial state (e.g., all hyperparameters set to zero). It selects a dataset uniformly at random.\\

Within each episode, the algorithm iterates over a maximum number of actions per episode ($T$). It updates the state ($s_t$) by obtaining the metafeatures of the dataset. It then selects actions while the current state is not terminal and not equal to the next state.\\

During action selection, the algorithm determines the next action ($a_t$). With probability $\epsilon$, it selects a random action uniformly from the hyperparameter grid ($\Lambda$); otherwise, it selects the action that maximizes the estimated Q-value ($\hat{Q}$).\\

The algorithm receives a reward ($r_t$) based on the loss function of the model trained with the selected hyperparameters. It generates the next state ($s_{t+1}$) based on the current state, selected action, and received reward. If the current state is equal to the next state, the algorithm terminates the current episode.\\

The algorithm stores the experience tuple (state, next state, action, reward) in the replay buffer ($B$). It manages the replay buffer by removing the oldest experience if its size exceeds a predefined size ($N_b$).\\

It then samples a mini-batch of experiences from the replay buffer and updates the parameters of the Q-network ($\theta$) by minimizing the mean squared error between the predicted Q-values ($\hat{Q}$) and the target Q-values ($\hat{Q}_{\text{target}}$).\\

The algorithm periodically updates the target Q-network parameters ($\theta^-$) to match the current Q-network parameters ($\theta$). Finally, it returns the updated Q-network parameters ($\theta$).\\

\begin{algorithm}[]
\caption{Find Optimal Hyperparameters}
\begin{algorithmic}[1]
\REQUIRE $q\_network$: Trained $\hat{Q}$-network,
         $\theta$: Trained $\hat{Q}$-network parameters,
         $D$: Dataset,
         $\Lambda$: Hyperparameter grid.
\STATE $\lambda^{*} \leftarrow$ None
\STATE $Q^{*} \leftarrow -\infty$
\STATE $s_t$ = (metafeatures($D$), ($\{0\}^{\text{dim}(\Lambda)},0$))
\FOR{$i \in 1,...,N$}
    \FOR{$a$ in $\Lambda$}
        \STATE $q\_value \leftarrow q\_network.predict(s_t, a; \theta)$
        \IF{$q\_value > best\_q\_value$}
            \STATE $\lambda^{*} \leftarrow a$
            \STATE $Q^{*} \leftarrow q\_value$
        \ENDIF
    \ENDFOR
    \STATE $s_{t+1} = \tau(s_t,\lambda^{*},r_{\lambda^{*}})$
    \STATE Terminate; if $s_{t+1} == s_{t}$
\ENDFOR
\RETURN $\lambda^{*}$
\end{algorithmic}
\label{algo:Hyp-RLfindopt}
\end{algorithm}

After getting the parameters of the $\hat{Q}$ network from the Hyp-RL shown in Algorithm \ref{algo:Hyp-RL}, we need to find the optimal hyperparameter setting for any dataset for the specified model. Algorithm \ref{algo:Hyp-RLfindopt} shows the strategy to find optimal hyperparameters using a trained Q-network. Algorithm \ref{algo:Hyp-RLfindopt} is not present in the work of H.S. Jomma et al \cite{jomaa2019hyp}, but it is designed based on intuition for better understanding. The algorithm begins by initializing the best hyperparameter configuration (\( \lambda^* \)) and the corresponding highest Q-value (\( Q^* \)) to default values. Then, for each dataset (\( D \)), it initializes the initial state (\( s_t \)) by extracting metafeatures from the dataset and setting the hyperparameter configuration to an initial state with all hyperparameters set to zero. Next, it iterates over a predefined number of iterations (\( N \)). Within each iteration, it iterates over all hyperparameters in the hyperparameter grid (\( \Lambda \)). For each hyperparameter, it predicts the Q-value (\( q\_value \)) using the trained Q-network and compares it with the current best Q-value (\( best\_q\_value \)). If the predicted Q-value is greater than the current best Q-value, it updates the best hyperparameter configuration and the corresponding best Q-value. After iterating over all hyperparameters, it generates the next state (\( s_{t+1} \)) based on the selected best hyperparameter configuration. If the next state is the same as the current state, indicating termination, the episode terminates. Finally, the algorithm returns the best hyperparameter configuration (\( \lambda^* \)).\\

\begin{algorithm}
\caption{Q-learning for Hyperparameter Optimization \cite{qi2023hyperparameter}}
\begin{algorithmic}[1]

\REQUIRE $M$: Neural network model, 
         $D$: reference dataset, 
         $N$: number of trials, 
         $S$: search space, 
         $\alpha$: learning rate, 
         $\gamma$: discount factor, 
         $\epsilon$: exploration rate 
\STATE Initialize action space $AS$
\STATE Initialize $Q(s, a)$ arbitrarily
\STATE Randomly select a hyperparameter configuration from $S$ and compute the validation metric for $M$ on $D$
\STATE $n \gets 0$
\WHILE{$n < N$}
    \STATE Set the current best hyperparameter configuration as the initial state $s_0$
    \WHILE{true}
        \STATE Choose action $a_t$ from state $s_t$ using $\epsilon$-greedy policy
        \STATE Take action $a_t$, observe reward $r_t$, and compute next state $s_{t+1}$
        \STATE Update $Q$:
        \[
        Q(s_t, a_t) \gets Q(s_t, a_t) + \alpha \left( r_t + \gamma \max_a Q(s_{t+1}, a) - Q(s_t, a_t) \right)
        \]
        \STATE Compute change $c$:
        \[
        c = \frac{|s_t - s_{t+1}|}{s_t}
        \]
        \STATE $s_t \gets s_{t+1}$
        \STATE $n \gets n + 1$
        \IF{$c > 0.01$ or $n \geq N$}
            \STATE \textbf{break}
        \ENDIF
    \ENDWHILE
\ENDWHILE
\RETURN optimal hyperparameter configuration with its evaluation metric
\end{algorithmic}
\label{algo:Q}
\end{algorithm}

The algorithm designed by Qi et al. \cite{qi2023hyperparameter} shown in Algorithm \ref{algo:Q} is a Q-learning approach for hyperparameter optimization. It takes as input a neural network model \( M \), a reference dataset \( D \), the number of trials \( N \), the search space \( S \), the learning rate \( \alpha \), the discount factor \( \gamma \), and the exploration rate \( \epsilon \). \\

First, it initializes the action space \( AS \) and the Q-values \( Q(s, a) \) arbitrarily. Then, it randomly selects a hyperparameter configuration from the search space \( S \) and computes the validation metric for the neural network model \( M \) on the reference dataset \( D \). \\

The algorithm iterates over trials, where in each trial, it sets the current best hyperparameter configuration as the initial state \( s_0 \). Within each trial, it continues to update the hyperparameter configuration until a certain condition is met. \\

At each step, it chooses an action \( a_t \) from the current state \( s_t \) using an \( \epsilon \)-greedy policy, takes the action, observes the reward \( r_t \), and computes the next state \( s_{t+1} \). It then updates the Q-value according to the Q-learning update rule. \\

Additionally, it computes a change \( c \) to determine if the hyperparameter configuration has significantly changed. If the change is below a threshold or the maximum number of trials is reached, the algorithm terminates. Finally, the algorithm returns the optimal hyperparameter configuration along with its evaluation metric.\\

\section{Discussion}
H.S. Jomaa et al. \cite{jomaa2019hyp} and Qi et al. approach the hyperparameter optimization problem using the Q-learning approach in their studies and experimental results show that their approaches can find the optimal or near-optimal hyperparameter configuration within a limited number of trials. They formulate the hyperparameter optimization problem as a sequential decision problem or Markov Decision Process (MDP) before applying Q-learning. When selecting the action, both works employ a $\epsilon - greedy$ policy, and their research lacks an explanation for why they pick this particular policy. A comparative analysis of several of the action policies is presented by Hasan et al. \cite{10.1007/978-3-031-30105-6_30}, who conclude that the Softmax Policy performs better for their problem than the $\epsilon - greedy$ policy. We may explore alternative strategies in the framework of Hyperparameter Optimization (HPO) to obtain the optimal outcome, even though the problem models of Hasan et al. and HPO differ.\\

Secondly, a discrete finite search space is used in the studies of H.S. Jomaa et al. \cite{jomaa2019hyp} and Qi et al., however the selection criteria of the hyperparameter search space is not clarified in these works. The HPO problem can also be formulated as a sequential decision problem for a continuous finite search space, in which the values of the hyperparameters may come from different distribution models provided the range of the values of each hyperparameter. Furthermore, rather than determining the optimal hyperparameter setting, we can learn a policy and utilize it to determine the optimal hyperparameter for any other dataset, as suggested by H.S. Jomaa et al. \cite{jomaa2019hyp}.

\section{Conclusion}
This paper presents possible approaches for the open-ended area Hyperparameter Optimization. All the methods do, however, have some restrictions, most of which arise from the computational cost of finding the optimal hyperparameter setting while having a finite number of trials. While surrogated models suffer from hypotheses in predicting the next hyperparameter setting to evaluate, random search suffers from randomness in determining optimal hyperparameter settings. The Hyperparameter Optimization problem is presented as a sequential decision problem, and a reinforcement learning approach is applied to address the problem to reduce the reliance on the hypothesis. This paper presents the studies conducted by Qi et al. \cite{qi2023hyperparameter} and H.S. Jomaa et al. \cite{jomaa2019hyp}. This paper focuses on how the HPO is formulated as a Markov Decision Problem and how Q-learning is applied to solve the problem. While Qi et al.'s method finds the optimal hyperparameter setting for a particular model and dataset, H.S. Jomaa et al.'s approach involves learning a policy before determining the best hyperparameter setting for a dataset so that the learned policy can be used to determine the best hyperparameter setting for any dataset. Furthermore, this work highlights several gaps in the existing literature and raises some research questions that need to be explored in future research.

\bibliographystyle{unsrtnat}
\bibliography{references}

\end{document}